\documentclass[journal,twoside,web]{ieeecolor}
\usepackage{generic}
\usepackage{cite}
\usepackage{amsmath,amssymb,amsfonts}
\usepackage{algorithmic}
\usepackage{graphicx}
\usepackage{algorithm,algorithmic}
\usepackage{hyperref}
\hypersetup{hidelinks=true}
\usepackage{textcomp}
\def\BibTeX{{\rm B\kern-.05em{\sc i\kern-.025em b}\kern-.08em
    T\kern-.1667em\lower.7ex\hbox{E}\kern-.125emX}}
\usepackage{url}
\usepackage{multirow} 
\usepackage{booktabs}
\usepackage{color}
\usepackage{dsfont}
\usepackage[table,xcdraw]{xcolor}
 \usepackage{comment}
\definecolor{newcolor}{rgb}{.8,.349,.1}
\usepackage[normalem]{ulem}

\newcommand{\reviewertwo}[1]{{\color{black} #1}}

\newcommand{\alg}{HIBMatch}
\newcommand\T{\rule{0pt}{2.4ex}}       
\newcommand\B{\rule[-1.2ex]{0pt}{0pt}} 

\begin{document}
\title{HIBMatch: Hypergraph Information Bottleneck for Semi-supervised Alzheimer’s Progression}
\author{Zhongying Deng\textsuperscript{*}, Shujun Wang\textsuperscript{*}, Angelica I Aviles-Rivero, Zoe Kourtzi, Carola-Bibiane Schönlieb, and for the Alzheimer's Disease Neuroimaging Initiative
\thanks{Z. Deng and S. Wang contributed equally to this work.}
\thanks{Z. Deng and C. Schönlieb are with the Department of Applied Mathematics and Theoretical Physics, University of Cambridge, Cambridge CB3 0WA, UK
(e-mail: \{zd294, cbs31\}@cam.ac.uk). }
\thanks{S. Wang is with the Department of Biomedical Engineering, The Hong Kong Polytechnic University. She is also with the Research Institute for Smart Ageing as well as the Research Institute for Artificial Intelligence of Things, The Hong Kong Polytechnic University, Hong Kong, China (e-mail: shu-jun.wang@polyu.edu.hk).}
\thanks{A. Aviles-Rivero is with the Yau Mathematical Sciences Centre, Tsinghua University, Beijing 100084, China (e-mail: aviles-rivero@mail.tsinghua.edu.cn).}
\thanks{Z. Kourtzi is with the Department of Psychology, University of Cambridge, Cambridge CB2 3EB, UK (e-mail: zk240@cam.ac.uk).}}

\maketitle

\begin{abstract}
Alzheimer's disease progression prediction is critical for patients with early Mild Cognitive Impairment (MCI) to enable timely intervention and improve their quality of life. While existing progression prediction techniques demonstrate potential with multimodal data, they are highly limited by their reliance on {labelled} data and fail to account for a key element of future progression prediction: not all features extracted at the current moment may be relevant for predicting progression several years later. To address these limitations in the literature, we design a novel semi-supervised multimodal learning hypergraph architecture, termed \alg{}, by harnessing hypergraph knowledge based on information bottleneck and consistency {regularisation} strategies. 
Firstly, our framework {utilises} hypergraphs to represent multimodal data, encompassing both imaging and non-imaging modalities. Secondly, to {harmonise} relevant information from the currently captured data for future MCI conversion prediction, we propose a Hypergraph Information Bottleneck (HIB) that discriminates against irrelevant information, thereby focusing exclusively on {harmonising} relevant information for future MCI conversion prediction. Thirdly, our method enforces consistency {regularisation} between the HIB and a discriminative classifier to enhance the robustness and {generalisation} capabilities of HIBMatch under both topological and feature perturbations. Finally, to fully exploit the unlabeled data, \alg{} incorporates a cross-modal contrastive loss for data efficiency. 
Extensive experiments on the Alzheimer's Disease Neuroimaging Initiative (ADNI) dataset demonstrate that our proposed HIBMatch framework surpasses existing state-of-the-art methods in Alzheimer's disease prognosis.

\end{abstract}


\section{Introduction}
\label{sec:introduction}
\IEEEPARstart{A}{lzheimer’s} disease (AD) is a progressive and irreversible neurodegenerative disorder that results in a high level of dependence~\cite{zeisel2020world}. The disease progresses through different stages, starting with Mild Cognitive Impairment (MCI), {characterised} by memory problems, and eventually leading to AD~\cite{gauthier2006mild}. However, not all patients diagnosed with MCI convert to Alzheimer’s disease, and some may remain stable or even revert to normal. This distinction between MCI converters and MCI non-converters highlights the need for accurate prognosis prediction~\cite{lu2018multiscale,wang2019ensemble}. Although there is no known cure for AD, the development of algorithmic techniques for progression prediction has attracted great interest from the medical community. Early treatment for MCI progression is crucial for improving patients’ quality of life, making the value of accurate prognosis prediction evident.

\begin{figure*}[htb]
\centerline{\includegraphics[trim=70 0 85 235, clip, width=0.95\textwidth]{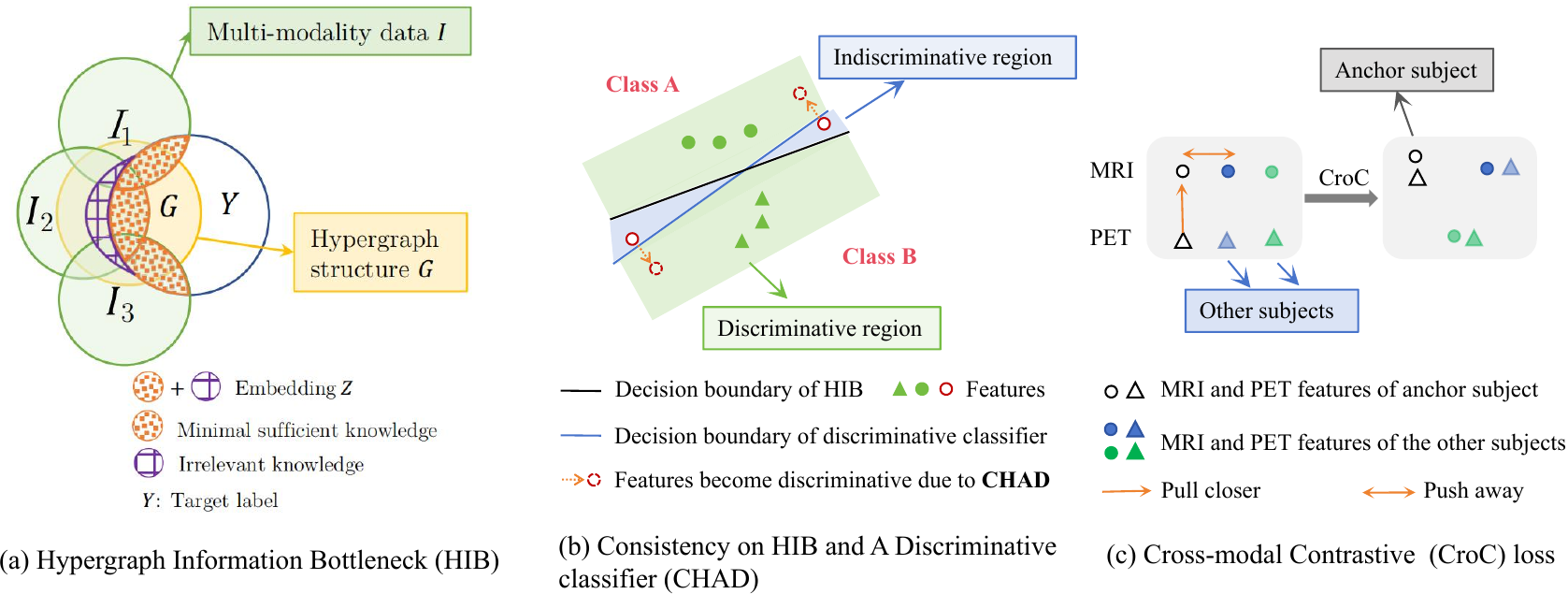}}
\vspace{-0.2cm}
\caption{(a) Hypergraph Information Bottleneck (HIB) {optimises} the representation $Z$ to capture the minimal sufficient information within the input data $D=(G,I)$ to predict the MCI conversion label $Y$. (b) Consistency on HIB and A Discriminative classifier (CHAD) updates the indiscriminative features, \textit{i.e.}, the red circles, which result in different predictions for HIB and a discriminative classifier, to be discriminative by forcing the features to achieve consistent predictions among two different classifiers.  (c) Cross-modal Contrastive (CroC) loss is applied to unlabelled data so that different modalities of the same subject are closer than the other subjects.}
\label{fig:principles}
\vspace{-0.5cm}
\end{figure*}

The literature has extensively investigated the task of predicting MCI to AD conversion from various perspectives. Early studies focused on analyzing a region of interest ~\cite{costafreda2011automated,davatzikos2011prediction,mcevoy2009alzheimer}, different single modalities or neuropsychological evaluation ~\cite{risacher2009baseline,qiu2009regional,fan2008structural}, and informative AD biomarkers ~\cite{hansson2006association,shaw2009cerebrospinal}. These have been widely explored using support vector machines (SVM) alone or in combination with Gaussian Radial functions ~\cite{toussaint2012resting,liu2022assessing}. With the advent of deep learning, researchers have explored the use of convolutional neural networks (CNNs) for MCI conversion prediction employing transfer learning and pyramid networks ~\cite{bae2021transfer,pan2020multi,shen2021heterogeneous,syaifullah2021machine,zhu2021long}.

Although there have been promising results in MCI conversion prediction in the literature, there are three major drawbacks that limit their performance. Firstly, existing techniques mainly rely on a single modality, which may not capture the full picture of the disease progression. To address this, recent studies have {emphasised} the need to develop multimodal techniques that can extract richer information ~\cite{aviles2022multi,grueso2021machine,shigemizu2020prognosis}. Secondly, how to effectively extract discriminative information from the multimodal data to facilitate conversion prediction and discard the irrelevant information from {highly heterogeneous} multimodal data has not been explored in the existing literature yet. Lastly, existing methods usually leverage a limited number of {labelled} data for model development whilst unlabelled data are not utilised~\cite{kang2023visual,cai2023discovering,pan2020multi}.  Unlabelled data are valuable because collecting the diagnosis results in future years for MCI conversion prediction is expensive and time-consuming. 

As a powerful tool to capture multimodal information, hypergraphs have been exploited for the diagnosis of AD~\cite{aviles2022multi,cai2023discovering}, but applying {them} for progression prediction remains unexplored. Hypergraphs generalise the concept of graphs by facilitating an edge connecting to more than two vertices. Therefore, compared with graphs that can only model pair-wise relations, hypergraphs excel in modelling high-order relations in multimodal information. As such, we leverage hypergraphs for progression prediction based on multimodal data, including Magnetic Resonance Imaging (MRI), Positron Emission Tomography (PET), and non-imaging information. 

Based on hypergraphs, this paper proposes a novel framework, namely the Hypergraph Information Bottleneck Matching the discriminative classifier (\alg). \alg{} integrates two effective strategies into the hypergraph framework to extract discriminative information from the multimodal data, i.e., information bottleneck~\cite{tishby2000information} and consistency {regularisation}. As shown in Fig.~\ref{fig:principles}(a), given the feature embedding, the Hypergraph-based Information Bottleneck (HIB) forces the hypergraph network to extract the discriminative information that is useful for predicting the MCI conversion labels, and meanwhile {discards} the irrelevant information from high heterogeneity multimodal data. \textbf{C}onsistency {regularisation} is applied between the \textbf{H}IB and \textbf{A} \textbf{D}iscriminative classifier (e.g., a fully connected layer followed by a SoftMax function), termed CHAD, which is illustrated in Fig.~\ref{fig:principles}(b). It updates the backbones to {enhance discriminative feature extraction} to achieve consistent predictions among HIB and the discriminative classifier. Finally, to leverage unlabelled images for further performance gain, we introduce a Cross-modal Contrastive (CroC) loss as depicted in Fig.~\ref{fig:principles}(c). This loss function pushes the features of different subjects away and meanwhile pulls different modalities of the same subjects closer. Thus, it can reduce the irrelevant information caused by the heterogeneity of multimodal data and draw the network's attention to the subject-level difference to better predict the progression of AD for each subject.

Our contributions are {summarised} as follows.
\begin{itemize}
    \item To the best of our knowledge, this is the first hypergraph framework specifically designed to leverage both multimodal and unlabelled data for progression prediction rather than diagnosis.
    \item We propose a Hypergraph Information Bottleneck (HIB) to represent the multimodal features and {harmonise} various modality data types. HIB integrates hypergraph and information bottleneck to learn discriminative representations that are meaningful with minimal but relevant information. 
    \item We devise consistency {regularisation} between the HIB and a discriminative classifier to further facilitate the discriminative feature extraction, thus supporting the robustness and {generalisation} capabilities of our method.
    \item We introduce a cross-modal contrastive loss to take advantage of unlabelled data for better performance, which is suitable for clinical practice where it is time-consuming to obtain the diagnosis results of future MCI conversion. 
    \item Experiments on the public ADNI dataset for AD prognosis prediction demonstrate that \alg{} outperforms other state-of-the-art methods, including hypergraph neural networks and semi-supervised learning frameworks. 
\end{itemize}

The rest of this paper is arranged as follows. We delve into related works in Section~\ref{sec:relatedwork}. We then elaborate on the technical details of our method in Section~\ref{method}. Experimental results and discussions are presented in Section~\ref{Experiments}. Finally, we {summarise} the conclusions of this paper in Section~\ref{conclusion}.

\section{Related Work}
\label{sec:relatedwork}
This section introduces previous studies on the progression prediction of Alzheimer's Disease, and other related machine learning concepts, such as graph and hypergraph learning, consistency {regularisation} as well as contrastive learning.

\subsection{Progression Prediction of Alzheimer's Disease}
Progression Prediction of AD at the Mild Cognitive Impairment (MCI) stage is crucial for the early treatment of potential AD patients. Existing methods implement the prediction either based on single modality or multimodality data. Single-modality-based methods use single-modality data, e.g., MRI or PET, which are usually simpler, whereas multimodality-based methods can {utilise} complementary information from multimodal data for better performance.

Single modality methods focus on designing new deep learning architectures to predict the progression of AD. 
Multiscale Deep Neural Network (MDNN)~\cite{lu2018multiscale} learns the patterns of AD pathology from Fluorodeoxy glucose PET (FDG-PET) using multi-scale feature vectors.
3D densely connected convolutional networks (3D-DenseNets)~\cite{wang2019ensemble} use multiple densely connected 3D networks~\cite{huang2017densely} with different hyperparameters as an ensemble to make robust predictions from MRI scans. 
Bae \textit{et al.}~\cite{bae2021transfer} transfer 3D convolutional networks, initially pre-trained for classifying normal control versus AD, to classify MCI converted to AD using MRI images. 
Multi-view Separable Pyramid Network (MiSePyNet)~\cite{pan2020multi} learns complementary information from axial, coronal, and sagittal views of PET using feature pyramid structures. 

Multimodal-based methods differ from each other in how to fuse multimodal information. Suk \textit{et al.}~\cite{suk2014hierarchical} employ a hidden layer to discover the shared information of MRI and PET images in a supervised manner. 
Spasov \textit{et al.}~\cite{spasov2019parameter} adopt a multimodal feature extractor to learn from structural MRI, demographic, neuropsychological, and APOe4 genetic data. Such a feature extractor concatenates features of different modalities and extracts useful information from them using modality-shared convolution or fully connected layers. 
Zhou \textit{et al.}~\cite{zhou2020multi} project MRI and PET modalities into a common latent space to reduce the heterogeneity of multimodal data. They further train multiple diverse classifiers and use the ensemble of these classifiers for more robust predictions. 
Li \textit{et al.}~\cite{li2020deep} {utilise} an autoencoder guided by mean discrepancy distance (MMD), which is a metric to measure distribution distance between different modalities, to integrate structural connectivity information of Diffusion Tensor Imaging (DTI) data into the resting-state functional MRI (rs-fMRI). 
These two {modalities of} data are also used in ~\cite{lei2020self} and their relevant information is learned using a low-rank regularizer to perform subspace learning. 
\reviewertwo{The cascaded multi-modal transformer~\cite{liu2023cascaded} leverages cross-attention to model pairwise modality interactions and can be generalised to incomplete data modality.} 
Latest works exploit graphs or hypergraphs to capture complex and high-order relationships among modalities, e.g., functional MRI (fMRI) and diffusion tensor imaging (DTI)~\cite{song2022multicenter}, arterial spin {labelling} MRI and blood-oxygen-level-dependent fMRI~\cite{li2019multimodal}. 
Our \alg{} also leverages hypergraphs but further improves them with information bottleneck which leads to a novel Hypergraph Information Bottleneck (HIB). Additionally, we propose a consistency {regularisation} on the HIB and a discriminative classifier to better extract discriminative features.

\subsection{Graph and Hypergraph Learning}
A graph usually comprises a vertex set and an edge set, with each vertex representing an entity and each edge modelling the pair-wise relationship between an entity pair~\cite{xia2021graph}. Exploiting these edges, a graph can capture the complex and relevant relationships among {vertices}. Graph learning further converts graph data into output vectors which can be regarded as the features of a graph. Then such features can be used for downstream tasks like vertex or entity classification. Since an edge of a graph can only model pair-wise relationships between {vertices}, it is less effective in capturing high-order relationships. To remedy this limitation, hypergraphs generalise graphs by enabling an edge connecting an arbitrary number of {vertices}~\cite {gao2020hypergraph}. Thus, hypergraphs can model even more complex relationships beyond pair-wise ones, and further extract high-order relationships, which widely exist in multimodal data. Due to this property, hypergraphs have been used to learn from the multimodal data for the diagnosis of AD.

Multimodal hypergraph diffusion network~\cite{aviles2022multi} leverages hypergraph to capture the high-order relationships among multimodal data, e.g., imaging and phenotypic data, for AD diagnosis. HSIA-GAN~\cite{bi2022hypergraph} integrates hypergraph and generative adversarial networks to capture both low-order and high-order information from imaging genetic data. 
Given multimodal phenotypes, HMGD~\cite{wang2023hypergraph} exploits a unified graph to model high-order similarity relationships among subjects and further introduces a hypergraph {regularisation} to select imaging phenotypes related to the risk single nucleotide polymorphism. 
BrainHGNN~\cite{cai2023discovering} proposes a brain network-tailored hypergraph neural network to capture the high-order interactions between multiple brain regions so that the propagation patterns of neuropathological events are identified for AD. Different from these hypergraph-based AD diagnosis methods, our \alg{} is designed for progression prediction of AD rather than diagnosis. \alg{} includes a novel Hypergraph Information Bottleneck (HIB), as well as a consistency {regularisation} on the HIB and a discriminative classifier.

\subsection{Consistency Regularization}

Consistency {regularisation} {and pseudo-labelling are} widely used for semi-supervised learning. \reviewertwo{Since pseudo-labelling, such as FixMatch~\cite{sohn2020fixmatch} and FlexMatch~\cite{zhang2021flexmatch}, adopts a threshold to select predictions from the model itself as the pseudo-labels, it may suffer from label noise and limited data which are common in medical imaging tasks. As such, we focus on consistency regularisation in this section. Its} key idea is that the network should output consistent predictions for the same unlabelled data under different perturbations. The perturbations include different image-level augmentations, like RandAugment~\cite{cubuk2020randaugment} and AutoAugment~\cite{cubuk2019autoaugment}, or feature-level perturbations, e.g., Dropout~\cite{srivastava2014dropout} and Dropblock~\cite{ghiasi2018dropblock}. These perturbations lead to different outputs, of which the distance is {minimised} to ensure consistent predictions. $\Pi$-Model~\cite{laine2016temporal} leverages a L2 distance to measure the consistency. Mean Teacher~\cite{tarvainen2017mean} introduces an exponential moving averaged model as a mean teacher to enforce consistency on a student model. Hereafter, most semi-supervised methods~\cite{berthelot2019mixmatch,berthelot2019remixmatch,sohn2020fixmatch,li2021comatch} stick to this mean teacher design. 
None of these methods {uses} a hypergraph information bottleneck (HIB) as a classifier to enforce consistency. Therefore, our method differs from them in enforcing consistency on the HIB and a discriminative classifier, with the HIB used as a new classifier for MCI progression prediction. 

\subsection{Contrastive Learning}
Contrastive learning is extensively used in self-supervised learning to learn instance discrimination from unlabelled data. It pulls different augmentations of the same image close so that they have similar feature embeddings, and pushes different images far away for instance discrimination. The well-known contrastive learning methods include SimCLR~\cite{chen2020simple}, SimCLRv2~\cite{chen2020big}, MoCo~\cite{he2020momentum} and MoCo v2~\cite{chen2020improved}. These methods provide powerful pre-trained models that can be fine-tuned on limited labels to obtain an impressive performance on various tasks. Our \alg{} does not focus on the pre-trained models, but seeks to apply contrastive learning to take advantage of unlabelled data for the progression prediction of AD. To reduce the irrelevant information caused by the heterogeneity of multimodal data, we further introduce a cross-modal design to contrastive learning.

\begin{figure*}[!t]
\centerline{\includegraphics[trim=20 2 245 3, clip, width=0.91\textwidth]{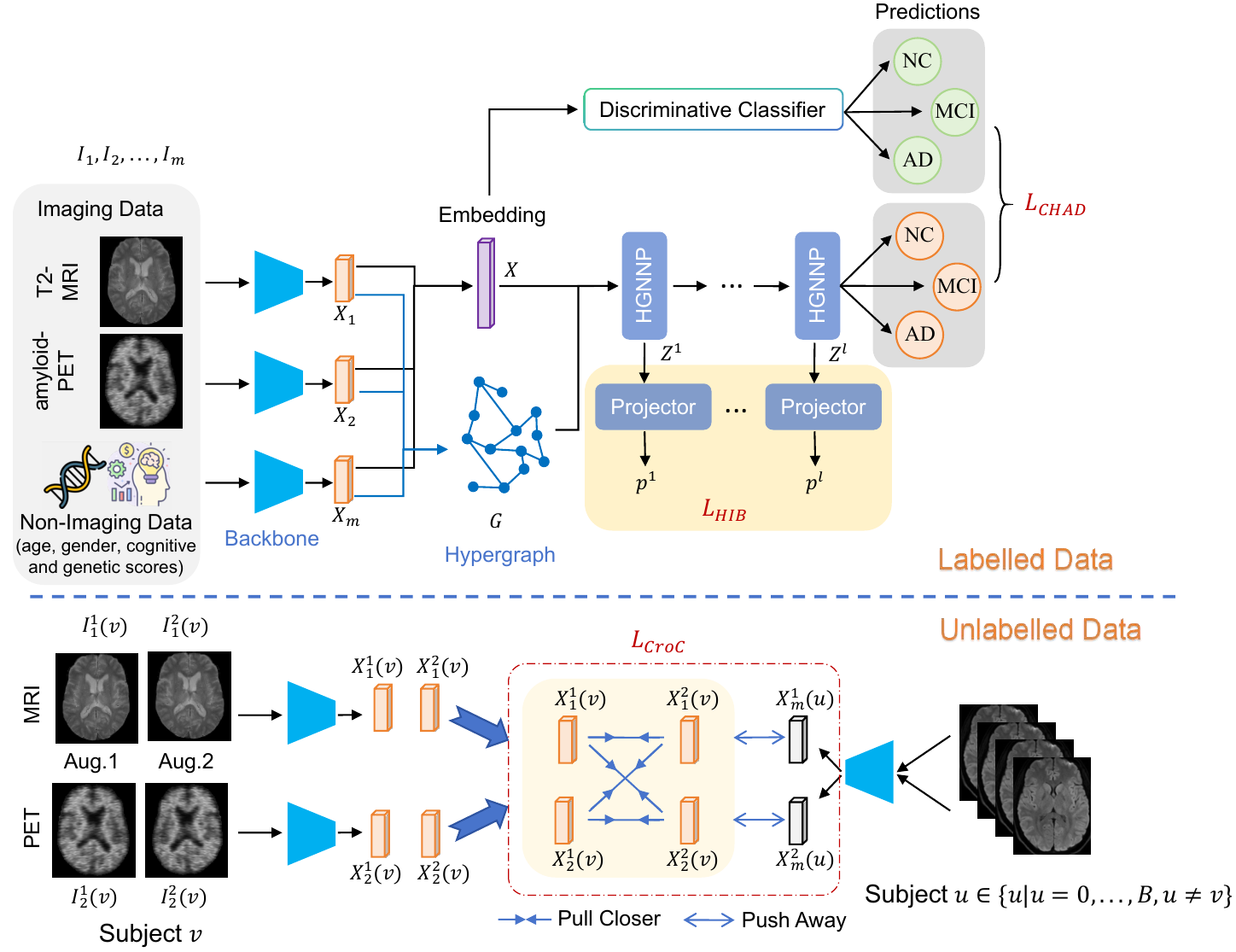}}
\vspace{-0.2cm}
\caption{Illustration of the whole workflow for labelled data (top) and unlabelled data (bottom). For labelled data, different modalities, including \reviewertwo{T2*-weighted MRI}, amyloid-PET, and non-imaging (age, gender, education years, cognitive and genetic scores), are input to different backbones for feature extraction. These features are used to construct a hypergraph which is then integrated with information bottleneck to obtain HIB. HIB is supervised by a HIB loss, $L_{HIB}$. HGNNP is a kind of hypergraph convolutional layer. The HIB and a discriminative classifier, trained using feature embeddings, are trained to predict the progression of AD. We further enforce a consistent prediction on their predictions by implementing a $L_{CHAD}$. For unlabelled data, the MRI and PET images of subject $v$ are augmented twice using different augmentation strategies, leading to four versions of feature representations. Then, the four versions are pulled closer and pushed far away from the features of the other subjects $u$ in a mini-batch of $B$ subjects. This is achieved by Cross-modal Contrastive (CroC) loss, $L_{CroC}$. {Note that only subjects with all three modalities are retained.}}
\label{fig:pipeline}
\vspace{-0.5cm}
\end{figure*}

\section{Proposed Framework} \label{method}

This section presents a detailed description of the crucial components of our proposed framework. The overview of the proposed method is illustrated in Fig.~\ref{fig:pipeline}. Firstly, we elucidate the process of constructing the hypergraph from a given set of multimodal data, and we delve into the specifics of the hypergraph convolution definition. Secondly, we explicate the fundamental principle of information bottleneck and integrate it into hypergraph neural networks, which leads to Hypergraph Information Bottleneck (HIB). Then, we enforce consistency on the HIB and a discriminative classifier as a {regularisation}. Lastly, we introduce a Cross-modal Contrastive (CroC) loss to leverage unlabelled data for model training. 

Our method uses human subjects from the Alzheimer’s Disease Neuroimaging Initiative (ADNI) datasets\footnote{Data used in the preparation of this article were obtained from the Alzheimer's Disease Neuroimaging Initiative (ADNI) dataset (\url{adni.loni.usc.edu}).}~\cite{jack2008alzheimer}, we confirm that informed consent was obtained from ADNI.

\subsection{Hypergraph Modelling and Hypergraph Convolution}

To overcome the
challenges associated with multimodality data ($I_1, I_2, ..., I_m$), we leverage a hypergraph
structure $G$ to represent the multimodal features ($X_1, X_2, ..., X_m$) extracted from backbones. Subsequently, we employ a hypergraph neural network to predict MCI conversion.

\textbf{Hypergraph representation learning.}
We consider an undirected attributed hypergraph $G = (V, E, \textbf{H})$ with a vertex set $V$, a hyperedge set $E$, and an adjacency matrix $\textbf{H} \in \mathbb{R}^{|V| \times |E|}$ for hyperedge weight. 
In our hypergraph structure, each vertex corresponds to a patient and hyperedges represent the relationship among different vertices. 
Unlike in a graph structure, a hyperedge in a hypergraph connects multiple vertices instead of just two, allowing for the representation of higher-order relationships, effectively grouping together subsets of vertices that share common features or properties.
Specifically, {the hyperedge weight between vertex $v$ and hyperedge $e$ can be defined as} 
$h_{v,e}= 
\left\{ 
    \begin{array}{lc}
        1, \ &\text{if} \  v \in e \\
        0, \ & \text{otherwise}
    \end{array}
\right.
$, {which are fixed during training.}
Moreover, vertex attributes can be seen as a feature embedding and denoted as $X$. 
Then, the input data can be represented as $D=(G, X)$. Therefore, under the multi-modal setting, we assume $m$ modalities as input and denote them as $D=(G, (X_1, X_2, ..., X_m))$.

A crucial step in hypergraph learning is the construction of the hypergraph structure. To achieve this, we first obtain the feature embeddings $X=\{X_1, X_2, ..., X_m\}$ from the multimodality data using a pre-trained network backbone, as illustrated in Fig.~\ref{fig:pipeline}. We then employ a {neighbour} strategy in feature space to generate the hyperedge groups following the same protocol as described in \cite{gao2022hgnn}. Specifically, for each vertex, its $k$-nearest {neighbours} in the feature space are connected by a hyperedge, resulting in the set of hyperedges $E_k=\{N_{\text{KNN}_k}(v)|v \in V\}$. 
These hyperedge groups are concatenated together to form a hypergraph for each modality {of} data. {To effectively utilize the multimodality knowledge, we concatenate $k$ different hypergraphs to generate the final hypergraph by $H=H_1\|H_2\|...\|H_k$. Then, we feed the resulting data $D$ into a Hypergraph Convolution Layer for further computation.}

\textbf{Hypergraph Convolution.}
We use spatial hypergraph convolution layers \cite{gao2022hgnn} for message aggregation. Messages can be passed either from vertex to hyperedge or from hyperedge to vertex using hyperpaths $P$, which is defined as $P(v_1,v_k) = (v_1,e_1,v_2,...,e_{k-1}, v_k)$. 
We define the inter-{neighbour} relation $N$ of hypergraph $G$ as $N=\{ (v,e) | w_{v,e}=1, v \in V, \text{ and } e \in E \}$. The vertex inter-{neighbour} set of hyperedge $e$ is defined as $N_v(e)=\{v|vNe\}$, and the hyperedge inter-{neighbour} set of vertex $v$ is defined as $N_e(v)=\{e|vNe\}$. To update the vertex information, we aggregate the messages from its hyperedge inter-{neighbours} $N_e(v)$, and to update the hyperedge information, we use the vertex inter-{neighbours} $N_v(e)$.
Thus, the spatial hypergraph convolution layer is defined as
\begin{equation}
\label{eq:feature_HIB}
f_e= \sum_{v \in N_v(e)} h_{v,e}  \cdot \frac{x_v}{|N_v(e)|}, \ \ \ \
{f}'_v=\sigma \Bigg(\sum_{e \in N_e(v)} \frac{f_e}{|N_e(v)|} \cdot \Theta \Bigg),
\end{equation}
where $x_v$, $f_e$, and ${f}'_v$ are the input, hidden, and output feature vectors. $\Theta$ is a trainable parameter of the current hypergraph convolution layer. $\sigma$ is a non-linear activation function, such as ReLU. The two-step message aggregation {realises} a closed message passing loop among vertices, which enables the model to capture higher-order relationships between the vertices in the hypergraph.

\subsection{Hypergraph Information Bottleneck (HIB)}
To balance the expressiveness and robustness of the model, we aim to {optimise} the vertex representation to capture the minimal sufficient information required for downstream tasks via the information bottleneck approach \cite{tishby2000information}. The Hypergraph Information Bottleneck (HIB) approach, as shown in Fig. \ref{fig:principles}(a), is derived from the Graph Information Bottleneck \cite{wu2020graph}, which requires the node representation $Z_v$ to minimize the information from hypergraph-structured data $D$ while maximizing the information to prediction $Y$.

 However, a major challenge in {realising} HIB numerically is the assumption of independent and identically distributed (IID) vertices, which is not feasible for many real-world scenarios. Therefore, we rely on a local dependence assumption for hypergraph-structured data, whereby given data related to a limited number of {neighbours} of vertex $v$, the data in the rest of the hypergraph is independent of $v$. 
 We assume a Markovian dependence to obtain the optimal representation, whereby the representation of each vertex is updated by incorporating its {neighbours} with respect to the hypergraph representation $X$. 
The information bottleneck seeks to optimise
\begin{equation}
\label{eq:hib}
\underset{\mathbb{P}(Z^l|D) \in \Omega }{min} \mathcal{L}_{\text{HIB}}(X,Y;Z^l):= [-I(Y;Z^l)+\beta I(X;Z^l)],
\end{equation}
where $\Omega$ characterises the space of conditional distribution of $Z^l$ given data $D$, and $\beta$ is a balancing weight. $l$ represents the $l$-th hypergraph convolution layer. 

We now define the mutual information $I(X;Z^{l})$,  between the initial vertex embedding and updated vertex embedding, following~\cite{nguyen2010estimating}. This yields the Cross-Entropy loss that reads:
\begin{equation}
\label{eq:celoss}
    I(Y;Z^l) \rightarrow -\sum_{v \in V} \text{CE}(Z_v^lW_{out};Y_v),
\end{equation}
where $W_{out}$ is the weight of the projector to predict the MCI conversion labels.

We then assume that the elements in $X^{l}$ are IID Bernoulli distributions: $Z^l = \bigcup_{i,j} \{ z_{i,j} \in  \{ 0,1  \} |z_{i,j}\overset{IID}{\sim } \text{Bernoulli(0.5)} \}$. So the mutual information can be defined as $I(X;Z^{l})=\frac{1}{nm} \sum_{i=1}^{n}\sum_{j=1}^{m} \text{KL}\left (\text{Bernoulli}\left (z_{ij}^l\right)\| \text{Bernoulli}\left (0.5\right )\right )$. Here, $\text{KL}$ denotes the Kullback-Leibler divergence between two Bernoulli distributions. 

{The choice of a Bernoulli(0.5) prior is grounded in fundamental information theory. Specifically, Bernoulli(0.5) is the maximum entropy distribution over random variables, which means it is the least informative or most unstructured prior possible. This makes it a natural baseline for variational information bottleneck methods, as it encodes no prior bias toward any particular latent configuration. Minimising the KL divergence between the posterior \( \mathbb{P}(Z^l | X) \) and this uninformative prior effectively penalises the amount of structure or complexity injected into the latent representation. In other words, it discourages the model from encoding unnecessary details about the input, especially high-frequency or modality-specific artefacts that are unlikely to generalise. 
Crucially, this penalty induces compression in the representation space by pushing the posterior distribution toward high entropy, thereby reducing the total mutual information \( I(X; Z^l) \). This supports the information bottleneck objective of discarding irrelevant information. Any deviation from Bernoulli(0.5) (e.g., toward a sparse or peaked prior) would inject inductive bias into the latent space and reduce entropy, possibly favouring spurious correlations — a serious risk under low-sample conditions.}

In summary, given feature representation $X$ from a pre-trained backbone, HIB aims to use hypergraph and information bottleneck to capture {the} minimal sufficient information required for future MCI prediction tasks. It updates $W_{out}$ and the node representation $Z^l$, thus updating $\Theta$ as $Z^l$ is obtained from $\Theta$ according to Eq. (\ref{eq:feature_HIB}). HIB does not update $X$ because our implementation of hypergraphs is based on DHG\footnote{DHG (\url{https://deephypergraph.com}) changes PyTorch tensors of features  $X$ to Numpy tensors, thus losing the gradient information of $X$.}, preventing the gradient of $\mathcal{L}_{HIB}$ from backpropagating to $X$. To enforce $X$ to be more discriminative, we propose CHAD (Sec.~\ref{sec:CHAD}) to update the backbone to output discriminative features $X$.

\subsection{Consistency between HIB and a Discriminative Classifier (CHAD)}
\label{sec:CHAD}
We expect that features $X$ should be sufficiently discriminative to achieve consistent predictions from HIB and the discriminative classifier, as illustrated in Fig.~\ref{fig:principles}(b). Therefore, we introduce a CHAD loss to promote discriminative feature learning: 
\begin{equation}
    \mathcal{L}_{CHAD} =  \frac{1}{B}\sum_v^B||\sigma(Z_v^lW_{out}) - \sigma(XW_{dis}) ||_2^2,
\end{equation}
where $B$ is batch size, $\sigma$ is a SoftMax function, and $Z_v^lW_{out}$ is the prediction of the HIB as defined in Eq.~\eqref{eq:celoss}. $X$ is the features and $W_{dis}$ is the weights of the discriminative classifier. It is worth noting that {optimising} $\mathcal{L}_{CHAD}$ will update the features $X$ by {optimising} the parameters of the backbones so that the HIB and the discriminative classifier can predict the same MCI conversion labels. The {underlying} assumption is that the features achieving consensus among different classifiers are more robust and {generalisable}, as illustrated in Fig.~\ref{fig:principles}(b).

\subsection{Cross-modal Contrastive Loss}
\label{sec:croc}
To leverage unlabelled image data, we further introduce a Cross-modal Contrastive (CroC) loss, as demonstrated in Fig.~\ref{fig:principles}(c). \reviewertwo{Here, unlabeled subjects refer to MCI cases without sufficient follow-up outcomes to determine whether they progress to AD or remain stable.} The CroC loss pulls the feature representations $X$ of different modalities of the same subjects closer to reduce the irrelevant information caused by the heterogeneity of multimodal data. Meanwhile, it pushes the features of different subjects away to facilitate the subject-level progression prediction of AD. Concretely, let us denote a batch of subjects as $\{I_1(v), I_2(v)\}_{v=0}^B$, with $I_1(v)$ and $I_2(v)$ being the MRI and PET multi-modality images of the subject $v$, respectively. Specifically, we apply different augmentations to both $I_1(v)$ and $I_2(v)$, leading to four images for the same subject, i.e., $I_1^1(v), I_1^2(v), I_2^1(v), I_2^2(v)$ as shown in Fig.~\ref{fig:pipeline}. These images are input to the backbones to obtain their corresponding features $X_1^1(v), X_1^2(v), X_2^1(v), X_2^2(v)$. A similar operation is also applied to the remaining subjects. Then the CroC loss is
\begin{equation}
\begin{aligned}
\label{eq:croc_loss}
    \mathcal{L}_{CroC} &=  -\frac{1}{B}\sum_v^B \Bigg\{ \lambda \bigg[ \log \frac{\exp(\cos(X_1^1(v), X_2^2(v))/\tau)}{\sum_{j, j\neq v}^B\exp(\cos(X_1^1(v), X_2^2(j))/\tau)} \\
    &+ \log \frac{\exp(\cos(X_2^1(v), X_1^2(v))/\tau)}{\sum_{j, j\neq v}^B\exp(\cos(X_2^1(v), X_1^2(j))/\tau)} \bigg] \\
    &+ \log \frac{\exp(\cos(X_1^1(v), X_1^2(v))/\tau)}{\sum_{j, j\neq v}^B\exp(\cos(X_1^1(v), X_1^2(j))/\tau)} \\
    &+ \log \frac{\exp(\cos(X_2^1(v), X_2^2(v))/\tau)}{\sum_{j, j\neq v}^B\exp(\cos(X_2^1(v), X_2^2(j))/\tau)} \Bigg\},
\end{aligned}
\end{equation}
where $\lambda=0.1$ is a trade-off hyper parameter and $\tau=0.5$ is the temperature parameter. The first $\log$ term in CroC loss {minimises} the distance between MRI and PET features of the same subject $v$, i.e., $X_1^1(v), X_2^2(v)$, and meanwhile enlarges the distance of different subjects $v$ and $j$. The second term plays a similar role. The last two terms pull different augmentations of the same subject closer, e.g., $X_1^1(v), X_1^2(v)$, and push different subjects away.

\subsection{Optimisation Scheme}

Our main task is MCI prediction conversion. This problem is taken from the perspective of a three-class classification task (NC, MCI, and AD). Cross-entropy loss is a common classification loss. However, in the medical domain, it is usual to encounter the class imbalance problem. To address this issue, we incorporate a focal loss instead of plain cross-entropy loss. We then define our overall optimisation scheme as
\begin{equation}
\label{eq:loss_all}
\begin{aligned}
    \mathcal{L}_{total} = &\frac{1}{|V|} \sum_{v \in V}\Bigl\{\text{CE}(P_v;Y_v) + \mu \left [-\alpha (1-P_v)^\gamma \text{log}(P_v) \right ] \Bigl\} \\
    &  + \beta (\mathcal{L}_{CHAD}  + \mathcal{L}_{CroC}) + \xi \frac{1}{L}\sum_{l=1}^{L} \mathcal{L}_{\text{HIB}},
\end{aligned}
\end{equation}
where $\mu, \beta$ and $\xi$ are balancing parameters. $\alpha$ and $\gamma$ are two hyper-parameters of the focal loss~\cite{lin2017focal} in our experiments, we set their values to 2 and 0.5, respectively.

\subsection{Disentangling Modality-Specific Noise from Clinically Relevant Variance}
{
To enhance the robustness of the learned representation $Z^l$, we aim to ensure that it captures clinically meaningful variance while suppressing irrelevant, modality-specific noise. Our design is based on the Information Bottleneck principle \cite{tishby2000information,wu2020graph}, formulated in Eq.~(\ref{eq:hib}), where $I(X; Z^l)$ is approximated via the KL divergence from a Bernoulli prior. This prior represents the maximum-entropy distribution and therefore encodes no prior structural assumption about the latent space, penalising any excess retention of input information that is not useful for prediction. This promotes sparse, low-information representations that retain only what is necessary for predicting $Y$, as illustrated in Figure~\ref{fig:principles}(a). }

{
Furthermore, the hypergraph convolution operation defined in Eq.~(\ref{eq:feature_HIB}) plays a crucial role in this disentanglement. By aggregating features over high-order relations among clinically similar subjects, it ensures that node representations reflect shared, stable patterns rather than modality-specific fluctuations. Subjects can belong to multiple hyperedges, allowing overlapping clinical and phenotypic similarities to be captured and propagated across the network. This overlapping structure is particularly effective in smoothing out noise and reinforcing clinically relevant signals through repeated exposure across different hyperedge contexts.}

{
To further disentangle useful signals from modality-specific artefacts, we incorporate consistency and contrastive regularisation. The CHAD loss enforces prediction alignment between the HIB and a discriminative classifier, driving the latent space toward modality-invariant, decision-consistent representations. The CroC loss ensures that embeddings of the same subject across different modalities are pulled together, while embeddings of different subjects are pushed apart. This structure-aware, information-theoretic strategy ensures that the latent space emphasises cross-modal, clinically relevant information and suppresses individual modality noise.
This integrated formulation allows HIB to disentangle shared diagnostic signals from unshared modality noise in a theoretically grounded and empirically validated manner.
}

{Though HIB, CHAD, and CroC build upon established ideas, such as the information bottleneck principle, consistency regularisation, and contrastive learning, our intent is to demonstrate how their targeted integration within a unified framework yields synergistic benefits for semi-supervised Alzheimer's progression.
Together, these components complement one another, leading to more robust multimodal learning under a semi-supervised setting. Our experiments validate that this targeted combination favourably outperforms each component in isolation, as well as several state-of-the-art baselines.}

\section{Experiments}
\label{Experiments}

\begin{table} [!t]
    \centering
    \caption{Demographics of subjects. Subjects suffering from MCI and converting to AD are also called MCIc or progressive MCI (pMCI) while  MCI non-converting to AD are known as MCInc.}
    \label{tab:demographic}
    \setlength{\tabcolsep}{0.5mm}{
    \begin{tabular}{l|cc|c|c}
        \toprule[1pt]
        \textbf{Subjects} & \multicolumn{2}{c|}{\textbf{MCInc}} & \textbf{MCIc/pMCI} &\multirow{2}{*}{\textbf{{Unlabelled}}}\\
        \cline{1-4}
        Progression to &  \textbf{CN} & \textbf{MCI} &\textbf{AD} & \\
        \toprule[1pt]
        Number       & 58 & 146 & 44 & 260\\
        Gender (M/F) & 17/41 & 77/69 & 19/25 & 148/112\\
        Age          &70.80$\pm$5.26 &71.43$\pm$7.03 & 72.14$\pm$7.41 & 74.19$\pm$6.68\\
        Educat. years  & 17.00$\pm$2.43 & 16.21$\pm$2.64 & 15.77$\pm$2.52 & 16.47$\pm$2.52\\
        \toprule[1pt]
    \end{tabular}
    }
    \vspace{-0.5cm}
\end{table}

In this section, we comprehensively introduce the involved dataset,  experimental settings, and experimental comparison results to validate our proposed \alg{} framework for prognosis prediction of Alzheimer's Disease.

\subsection{Dataset Description}
We evaluate the proposed \alg{} framework on the Alzheimer's Disease Neuroimaging Initiative (ADNI) dataset, which is one of the most popular datasets for Alzheimer's Disease. 
ADNI is a longitudinal multi-{centre} and multi-modality study designed for the early detection and tracking of Alzheimer’s disease. 
The dataset contains four categories: Normal Control (NC), early Mild Cognitive Impairment (EMCI), late Mild Cognitive Impairment (LMCI), and Alzheimer’s disease (AD) at the Baseline Visit while only three categories in the follow-up: NC, Mild Cognitive Impairment (MCI), and AD.
To explore the disease progression, we focus on patients diagnosed with MCI (EMCI/LMCI) at the baseline visit.
The goal is to forecast MCI conversion after a fixed two-year window to identify whether subjects converted to NC (probably misdiagnosis in the beginning) and AD or kept MCI, based on the multi-modality data, including \reviewertwo{T2*-weighted MRI, Amyloid-PET}, and non-imaging clinical information. \reviewertwo{The choice of T2*-weighted MRI was motivated by its sensitivity to tissue microstructure, iron deposition, and microvascular changes—biomarkers increasingly linked to Alzheimer’s disease. This modality complements Amyloid PET, which reflects amyloid burden, and non-imaging clinical variables capturing demographic and genetic risks, enabling the model to learn a more comprehensive representation of Alzheimer’s pathology.} 
In some literature~\cite{li2023early,pan2020multi},  those suffering from MCI and converting to AD are also called MCIc or progressive MCI (pMCI) while those from MCI non-converting to AD are known as MCInc. 
The Non-imaging information consists of demographic, genetic, and cognitive features, e.g., age, gender, education, APOE4, MMSE, ADNI-MEM \cite{crane2012development}, ADNI-EF \cite{gibbons2012composite}.
{For non-imaging data, we handle missing values using standard imputation techniques, such as mean or mode imputation, depending on the data type, to ensure compatibility across samples. However, for imaging modalities like MRI and PET, we do not currently perform modality imputation due to the high dimensionality and complexity involved.} {Only subjects with all three modalities were retained while those with incomplete modalities were filtered out.}
After the filter, we got 248 patients from ADNI-2 who had all three modalities at the Baseline Visit and diagnosis results two years later. We also used 260 patients with three modalities but {without 24-month outcomes} as unlabelled data for efficient learning. The demographics of these subjects are shown in Table~\ref{tab:demographic}. The {labelled} data occupies about 48.82\% of the total subjects (i.e., 248 out of 508 subjects).

\begin{table*}[tb]
    \centering
    \caption{Comparison of our method with state-of-the-art methods for the task of 1) MCIc vs. NC (MCI converting to NC) and 2) MCIc vs. MCI (unchanged over time). AUC (\%) and accuracy (ACC, \%) are reported and the top results are highlighted in bold. * denotes results adapted from~\cite{li2023early}.}
    \label{tab:AD_NC_SoTA}
    \setlength\tabcolsep{4pt}
    \begin{tabular}{c|c|c|cc}
    \toprule[1pt]
       \textbf{Methods} &\textbf{Data Type} & \textbf{Subjects} &  \textbf{AUC} & \textbf{ACC} \\
        \midrule
        Multimodal CNN~\cite{huang2019diagnosis} &MRI+PET & 441 MCInc, 326 pMCI/MCIc & 77.49 &72.22\\
        MiSePyNet~\cite{pan2020multi} & PET& 360 MCInc, 166 pMCI/MCIc &86.80 & 83.05 \\
        HGNN~\cite{feng2019hypergraph}* &MRI+Morphology &133 MCInc, 75 MCIc&  70.47 &66.72 \\
        DHGNN~\cite{jiang2019dynamic}* &MRI+Morphology &133 MCInc, 75 MCIc &  69.14 &66.73 \\
        HGNN+~\cite{gao2022hgnn}* &MRI+Morphology &133 MCInc, 75 MCIc & 70.62 &64.77 \\
        HAN~\cite{li2023early}* &MRI+Morphology &133 MCInc, 75 MCIc &76.19 &71.10 \\
        \hline
        \multirow{2}{*}{\alg{} (Ours)} &\multirow{2}{*}{MRI+PET+Non-Imaging} &\multirow{2}{*}{204 MCInc, 44 MCIc, \textbf{260 {Unlabelled}} } &\cellcolor[HTML]{D7FFD7}\textbf{95.23}\textsuperscript{1}  &\cellcolor[HTML]{D7FFD7}\textbf{94.59} \\
        & & & \cellcolor[HTML]{D7FFD7}\textbf{85.93}\textsuperscript{2} &\cellcolor[HTML]{D7FFD7}\textbf{87.68} \\
    \bottomrule
    \end{tabular}
     \\ \textsuperscript{1} The results of AD (MCIc) vs. NC, and \textsuperscript{2} AD (MCIc) vs. MCI.
     \vspace{-0.3cm}
\end{table*}

\begin{table*} [tb]
    \centering
    \caption{Comparison of our method (\alg{}) with existing hypergraph and semi-supervised techniques for classification task of 1) MCI progression to AD vs. Non-AD (i.e., MCIc vs. MCInc) and 2) MCI progression to AD vs. MCI vs. NC (i.e., pMCI vs. MCI subjects of MCInc vs. NC subjects of MCInc). All the methods use the same data type and subjects. 
    The top results are highlighted in \textbf{bold}. Average results (\%) are reported by five-fold cross-validation.}
    \label{tab:results-SOTA}
    {
        \setlength\tabcolsep{5pt}
        \begin{tabular}{l|ccc|ccc}
            \toprule[1pt]
            
             \multirow{2}{*}{\textbf{Methods}} & \multicolumn{3}{c|}{{AD vs. Non-AD (MCIc vs. MCInc)}} & \multicolumn{3}{c}{{AD (MCIc) vs. MCI vs. NC }}\\
             \cline{2-7}  & \textbf{AUC} & \textbf{PPV} & \textbf{NPV} & \textbf{AUC avg.} & \textbf{PPV avg.} & \textbf{NPV avg.}  \T \B \\
             \midrule
            \T
            \textbf{HGNN}~\cite{feng2019hypergraph} &83.28$\pm$8.39 &90.00$\pm$20.36 &84.94$\pm$9.72  & 76.69$\pm$5.66 &74.84$\pm$10.04  & 82.04$\pm$4.68  \\ 

            \textbf{HGNN+}~\cite{gao2022hgnn} & 81.39$\pm$4.36 & 82.00$\pm$40.25 & 83.99$\pm$10.68  & 75.75$\pm$2.94 &75.67$\pm$8.59  & 79.79$\pm$5.11   \\ 

            \hline
            \textbf{FixMatch}~\cite{sohn2020fixmatch} &89.21$\pm$8.14 & 83.16$\pm$8.86 & 87.18$\pm$8.61 &79.92$\pm$2.82 &74.29$\pm$8.86 &80.78$\pm$5.71\\
            \textbf{NorMatch}~\cite{deng2024normatch} &87.33$\pm$14.55 & \cellcolor[HTML]{D7FFD7}\textbf{94.55}$\pm$12.20 & 87.55$\pm$10.04 &78.39$\pm$6.28 &\cellcolor[HTML]{D7FFD7}\textbf{78.55}$\pm$7.40 &82.18$\pm$5.90\\ 
            \hline
            \textbf{\alg{}} (Ours) & \cellcolor[HTML]{D7FFD7}\textbf{91.70}$\pm$4.87 & 89.67$\pm$14.37 &\cellcolor[HTML]{D7FFD7} \textbf{90.02}$\pm$4.98  & \cellcolor[HTML]{D7FFD7}\textbf{80.71}$\pm$3.38 &76.76$\pm$7.82  & \cellcolor[HTML]{D7FFD7}\textbf{83.13}$\pm$5.91   \\ 

            \toprule[1pt]
        \end{tabular}
    }
    \vspace{-0.4cm}
\end{table*}

\begin{figure*}[tb]
\centering
\setlength{\abovecaptionskip}{0pt}
\setlength{\belowcaptionskip}{-2pt}
\includegraphics[width=0.90\linewidth]{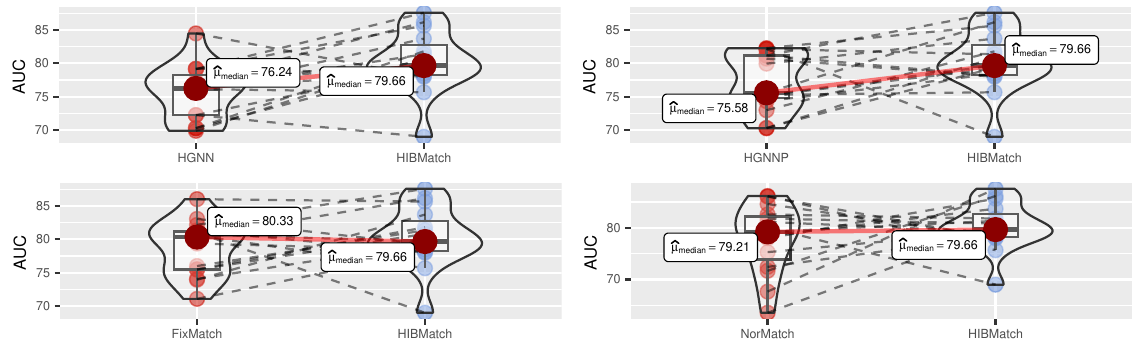}
\caption{Pairwise comparisons between HIBMatch and existing methods. Each subplot shows the AUC distribution for the paired methods, with dashed lines connecting individual samples. The red and blue circles indicate median AUC scores.}
\label{fig_b}
\vspace{-0.4cm}
\end{figure*}

\begin{figure}[tb]
\centering
\setlength{\abovecaptionskip}{0pt}
\setlength{\belowcaptionskip}{-2pt}
\includegraphics[width=1\linewidth]{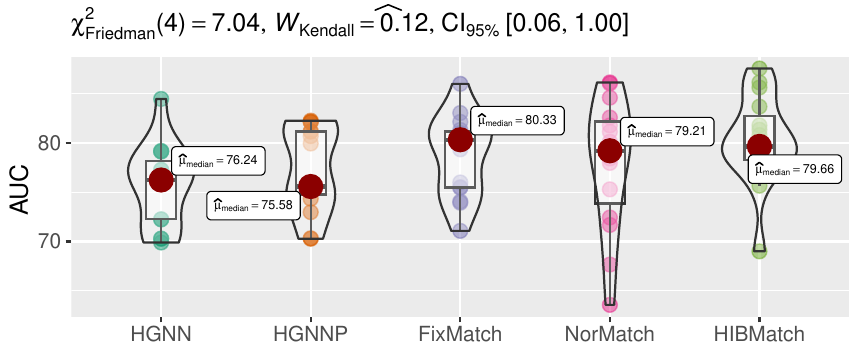}
\caption{Statistical Analysis: (a) Violin plots of AUC scores for each method, with the Friedman test results (\reviewertwo{$\chi^2 = 7.04, W = 0.12, \text{CI}_{95\%} [0.06, 1.00]$).}}
\label{figA}
\vspace{-0.4cm}
\end{figure}

\subsection{Dataset Pre-processing}
\label{sec:data_preprocess}
For the data pre-processing, all MRI volumes are processed following (1) anterior commissure (AC)-posterior commissure (PC) alignment, (2) skull stripping, (3) intensity correction, (4) cerebellum removal, and (5) linear alignment to a template MRI.
The corresponding PET volumes are aligned to their MRI volume via linear registration.
After that, the {labelled} images are randomly flipped and centre-cropped as augmentations. For unlabelled images, we further apply extra augmentation policies (as stated in Section~\ref{sec:croc}) which include random rotation and random adjusting contrast besides flipping and cropping. {To preserve diagnostic semantics, we restrict rotation angles to be small ($\pm$15$^{\circ}$) so that it is diagnostically tolerable. Moreover, contrast adjustments are bounded (with a gamma value in [0.6, 1.5]), simulating scanner-related variability rather than artificial distortion.} 
For the Non-imaging data, we {normalise} each feature in the range of [0,1].

\subsection{Training Details}
\label{sec:training_details}
We implemented our framework using PyTorch (\url{https://pytorch.org/}) and the DHG (\url{https://deephypergraph.com}) library on one NVIDIA A100 GPU. We use DenseNet-121~\cite{huang2017densely} as the backbone for PET/MRI images, and a single fully-connected layer as a backbone for non-image data. The hypergraph in our framework is HGNN+~\cite{gao2022hgnn}. {The discriminative classifier receives the same backbone features as the HIB branch. Specifically, the features of each modality have 1024 dimensions, and the three modalities are concatenated as 3072-dimensional features, which are then fed to the discriminative classifier.} The backbone models and the hypergraph are then trained with AdamW~\cite{loshchilov2017decoupled} optimiser with a batch size of 30. The learning rate was set as 1× 10$^{-3}$ and decreased to 0. We set the number of neighbours to 20 when constructing the hypergraph.
We empirically set the hyper-parameters $\mu,\beta$ and $\xi$ as 1, 0.1, and 10, respectively.

\subsection{Evaluation Protocol}
We discuss the following conditions in our experiments: (1) the classification accuracy of \alg{} compared with state-of-the-art hypergraph neural network methods as well as the state-of-the-art (SOTA) semi-supervised methods; (2) performance comparison under different ratios of {labelled} data of our method vs. existing SOTA techniques;
(3) robustness analysis under two kinds of attacks. 
We follow standard protocol in the medical domain and use the area under the ROC curve (AUC), positive predictive value (PPV \%), and negative predictive value (NPV \%) to measure the performance.
For the results, we report the average performance and standard deviation over five-fold cross-validation, where cross-validation is applied to {labelled} data with a ratio of 8:2 for training and validation.

\subsection{Results \& Discussion}
Below we show the results of MCI progression prediction, i.e., forecasting the conversion results of the subjects with MCI at the baseline visit to predict whether the subjects progress to NC, AD, or remain MCI two years later.
\subsubsection{Comparison to the State-of-the-Art Methods}
We begin our evaluation by comparing our framework against state-of-the-art multimodal methods, including those 
using vanilla CNN~\cite{huang2019diagnosis}, using CNN with multi-view feature fusion~\cite{pan2020multi} and using hypergraphs~\cite{feng2019hypergraph,jiang2019dynamic,gao2022hgnn,li2023early}. The task is to predict MCI conversion after a fixed time window to identify whether subjects converted to NC, AD, or remain unchanged. The results of 1) MCIc vs. NC subjects of MCInc and 2) MCIc vs. MCI subjects of MCInc are presented in Table~\ref{tab:AD_NC_SoTA}.

It is clear that our \alg{} achieves the best AUC and ACC. More importantly, owing to the use of {unlabelled} data, our \alg{} can beat the other multimodal methods leveraging {unlabelled} data. This is promising because the labels, i.e., diagnosis results, of AD in the future years are expensive to obtain. As such, leveraging {unlabelled} data for model development is essential for saving medical resources.

Then, we compare our \alg{} against two state-of-the-art hypergraph neural networks and two semi-supervised methods under the same setting, considering that different methods in Table~\ref{tab:AD_NC_SoTA} may use different data types and subjects, leading to an unfair comparison. 
Specifically, HGNN~\cite{feng2019hypergraph} is a hypergraph neural network based on spectral convolution. 
HGNN+~\cite{gao2022hgnn}, on the other hand, is an extended version of HGNN which is a general high-order multimodal data correlation {modelling} framework.
For semi-supervised learning, FixMatch~\cite{sohn2020fixmatch} is a well-known semi-supervised learning method that uses both consistency {regularisation} and pseudo-labelling to take advantage of massive unlabelled data. 
NorMatch~\cite{deng2024normatch} is the latest semi-supervised method that exploits {normalising} flows as a generative classifier to deal with the label noise in the pseudo-labelling strategy.

We report the quantitative comparison of our \alg{} vs. existing techniques in Table \ref{tab:results-SOTA}.
It is observed that our \alg{} framework reports the best AUC and NPV performance on both the binary-class task (i.e., separation of AD from MCI/NC cases in the next two years, given the MCI subjects at the baseline visit) and three-class one (predicting MCI conversion to AD vs. MCI vs. NC). Notably, for the binary-class task of AD vs. non-AD, our \alg{} achieves the best performance of 91.7\% for AUC, outperforming the second-best FixMatch by 2.49\%. The favourable performance of identifying these potential AD patients can facilitate early intervention. 
The highest NPV value of 90.02\% (binary-class) and 83.13\% (three-class) also indicates that \alg{} is more reliable for identifying true negative cases and avoiding false negatives. 
For PPV, our \alg{} is worse than NorMatch mainly because some MCI patients are mistakenly classified into AD. However, NorMatch {falls} short in the overall metric of AUC, i.e., 4.37\% and 2.32\% worse than \alg{} in binary-class and three-class tasks, respectively, suggesting that it struggles to trade off the true positive rate against the false positive rate. 
Across all the results, our framework demonstrates that utilising the information bottleneck can further improve the network progression prediction ability over these vanilla hypergraphs. Additionally, our semi-supervised design is also effective when compared with state-of-the-art semi-supervised methods. Note that the standard deviations of different methods can be large because the results are averaged over five-fold cross-validation.

To further support our experimental results, we conducted a series of statistical tests. First, a non-parametric Friedman test was used to assess multiple comparisons between the different methods. We employed Kendall's coefficient of concordance (WW) as a measure of effect size, with a 95\% confidence interval (CI). The results, shown in Figure~\ref{figA}, indicate a statistically significant difference in performance ($\chi^2 = 7.04, W = 0.12$). The effect size, $W_{Kendall}=0.12$, further supports the relevance of the differences observed. \reviewertwo{Instead of reporting DeLong confidence intervals, we use the AUC values per method obtained via repeated cross-validation to empirically capture performance variability. This approach aligns with our use of nonparametric tests and fold-level comparisons.
We conducted a post hoc power analysis to assess the statistical robustness of our findings. Based on the observed effect size (Kendall’s $W = 0.12$), the estimated statistical power was {22.35\%}. This limited power reflects the small number of independent observations used in the test (over 5-fold evaluation), not the size of the underlying dataset. Despite this, {HIBMatch consistently achieved the highest average and median AUC} across all folds and runs. These consistent results, even under conditions of high variance and limited sample granularity, underscore the robustness and practical effectiveness of our proposed method.}
Moreover, as highlighted in the plot, our proposed method, HIBMatch, achieved the highest median AUC score of 79.66, outperforming other methods like HGNN and HGNNP, which had median AUC scores of 76.24 and 75.58, respectively. This suggests that HIBMatch provides more robust and consistent performance compared to existing methods, demonstrating its effectiveness in achieving superior results.
To gain deeper insight into the comparative performance of our proposed method (HIBMatch) against existing techniques, we conducted pairwise comparisons using the Wilcoxon signed-rank test. The results are illustrated in the pairwise violin plots in Figure~\ref{fig_b}, which show the AUC distributions for each method compared to HIBMatch. The pairwise Wilcoxon test results indicate that HIBMatch achieves consistently higher performance than HGNN and HGNNP, and performs competitively against FixMatch and NorMatch. These results {emphasise} HIBMatch’s robustness and reliability across different comparisons.

\subsubsection{Performance Comparison over Different Label Counts}
By default, we have about 48.82\% total subjects with labels, i.e., 248 out of 508 subjects. To demonstrate the effectiveness of our proposed method with limited labelled data, we conducted experiments with four different label counts: 160, 120, 80, and 40 labelled subjects, corresponding to 31.50\%, 23.62\%, 15.75\%, 7.87\% of total subjects. This setting is of clinical significance because the labels (i.e., diagnosis results) take 24, 36, or even more months to be obtained, which is expensive. Fully leveraging the data collected at baseline visits, but without the labels, consuming several years to obtain, to predict MCI conversion can save clinical resources. 

The results of fewer label counts are reported in Table~\ref{tab:results-efficient}, which shows that \alg{} outperforms all compared techniques in terms of AUC. We focus on the AUC because it is an overall metric trading off the true positive rate (TPR) against the false positive rate (FPR). 
We observed that even when trained with fewer annotated data samples, our method still achieves promising performance despite a decrease compared to the full set of annotated data. 
This suggests that \alg{} can effectively learn from limited labelled data and {generalise} well to the remaining testing data by identifying relevant information (\textit{e.g.}, features and patterns), thus improving progression prediction.

  \begin{table} [tb]
    \centering
    \caption{AUC (\%) of our method (\alg{}) and existing hypergraph techniques under different training sample settings for the task of MCI progression to AD vs. MCI vs. NC. 
    The top results are highlighted in \textbf{bold}.}
    \label{tab:results-efficient}
    {
        \setlength{\tabcolsep}{0.8mm}{
        \begin{tabular}{l|c|c|c|c}
            \toprule[1pt]
            
             {\textbf{Methods}} & \textbf{31.50\% } &  \textbf{23.62\% } & \textbf{15.75\% } & \textbf{7.87\%} \\
             \hline
            \T
            
            {HGNN} & 72.99$\pm$2.65	&74.30$\pm$5.00 &72.75$\pm$7.37	& 71.50$\pm$10.84\\ 

            {HGNN+} & 73.22$\pm$4.74	&75.76$\pm$6.08		&74.62$\pm$6.40	& 72.92$\pm$9.40 \\ \hline
           {FixMatch} &79.93$\pm$4.51		&80.28$\pm$7.81		&73.53$\pm$3.57	& 68.73$\pm$5.39 \\
            {NorMatch} &76.20$\pm$5.53	&73.32$\pm$9.86		&70.87$\pm$4.08	&71.27$\pm$12.39 \\ \hline
            {\alg{}} & \cellcolor[HTML]{D7FFD7}\textbf{80.29}$\pm$6.80		&\cellcolor[HTML]{D7FFD7}\textbf{80.69}$\pm$4.32 	&\cellcolor[HTML]{D7FFD7}\textbf{74.71}$\pm$1.30	& \cellcolor[HTML]{D7FFD7}\textbf{73.90}$\pm$5.47\\ 

            \toprule[1pt]
        \end{tabular}
        }}
        \vspace{-0.4cm}
\end{table}

  \begin{table*} [tb]
    \centering
    \caption{{Robustness analysis for the task of MCI progression to AD vs. MCI vs. NC.
    We use two types of attacks: dropping hyperedges for topological attacks and injecting noise into embeddings for feature attacks. Noise means noise injection into the feature embedding. $\rho$ is the feature noise ratio for noise injection.
    The top results are highlighted in \textbf{bold} font.}}
    \label{tab:results-robust}
    {
        \setlength{\tabcolsep}{1mm}{
        \begin{tabular}{l|c||c|c|c||c|c|c}
            \toprule[1pt]
            \multirow{2}{*}{\textbf{Methods}} & \multirow{2}{*}{\textbf{No Attack}} & \multicolumn{3}{c||}{\textbf{Drop Hyperedges}} & \multicolumn{3}{c}{\textbf{Noise Injection}} \\
             \cline{3-8} 
            &  & 20\% & 40\% & 60\% &$\rho=0.01$ & $\rho=0.05$ & $\rho=0.1$
              \T \B \\
             \hline
            \T
            \textbf{HGNN+}~\cite{gao2022hgnn} & $75.75\pm2.94$ & $77.49\pm3.68$ &$72.79\pm3.40$ &$66.11\pm4.89$ & $78.72\pm2.90$ &$75.86\pm3.91$ & $75.20\pm3.79$  \\
             \textbf{\alg{}} (Ours) &\cellcolor[HTML]{D7FFD7} $\textbf{80.71}\pm 3.38$  &\cellcolor[HTML]{D7FFD7}$\textbf{79.52}\pm 2.91$  &\cellcolor[HTML]{D7FFD7}$\textbf{79.66}\pm5.43$ &\cellcolor[HTML]{D7FFD7}$\textbf{79.55}\pm2.58$ &\cellcolor[HTML]{D7FFD7}$\textbf{79.95}\pm 5.34$ &\cellcolor[HTML]{D7FFD7}$\textbf{79.24}\pm 4.29$  &\cellcolor[HTML]{D7FFD7}$\textbf{78.11}\pm 3.37$ \\ 
            \hline
            \toprule[1pt]
        \end{tabular}
        \vspace{-0.6cm}
    }}
\end{table*}

\subsubsection{Robustness Analysis under Attacks}
To study the robustness of the proposed \alg{} under the hypergraph topology and feature tasks, we conduct experiments by (1) randomly dropping 20\%, {40\%, and 60\%} hyperedges from the original hypergraph for structure attack; (2) randomly injecting feature noise to the extracted feature embedding $X$ for feature attack. 
Specifically, we define the mean of the maximum value of each vertex feature as $r$, random Gaussian noise as $\epsilon \sim  N(0,1)$, and feature noise ratio as $\rho$.
We then calculate noise according to $\eta=\rho \cdot r \cdot  \epsilon$. 
We set $\rho$ as 0.01, {0.05, and 0.1} in the experiments.
The results are shown in Table~\ref{tab:results-robust}. 
The results, shown in Table 3, demonstrate that \alg{} consistently outperforms the previous state-of-the-art HGNN+~\cite{gao2022hgnn} in both scenarios, indicating that \alg{} enhances the model's robustness for structure and feature perturbations. {Notably, when dropping 60\% hyperedges, HIBMatch beats the HGNNP+ by 13.44\%,} supporting its effectiveness for {generalisation} and robustness.

\begin{table} [tb]
    \caption{Ablation study for the task of MCI progression to AD vs. MCI vs. NC.  
    The top results are highlighted in \textbf{bold} font.}
    \label{tab:ablation}
    {
        \setlength{\tabcolsep}{0.5mm}{
        \begin{tabular}{l|c|c|c}
            \toprule[1pt]
            \textbf{Methods} & \textbf{AUC (\%)} & \textbf{PPV (\%)} & \textbf{NPV (\%)} 
              \T \B \\
             \hline
            \T
            Hypergraph only  & $69.69$ & $65.88$ & $79.44$\\
            Hypergraph Info. Bottl. (HIB) &  $72.36$ &$61.43$  & \cellcolor[HTML]{D7FFD7}$\textbf{84.74}$  \\
            HIB+CHAD & $79.23$ & $76.13$ & $81.16$   \\  
            HIB+CHAD+CroC (\alg{}) & $\cellcolor[HTML]{D7FFD7}\textbf{80.71}$ & \cellcolor[HTML]{D7FFD7}$\textbf{76.76}$ & $83.13$   \\ 
            \hline
            \toprule[1pt]
        \end{tabular}
    }}
    \vspace{-0.4cm}
\end{table}

\subsubsection{Ablation Study}
\textbf{Effectiveness of HIB.} We validate the effectiveness of introducing an information bottleneck to the hypergraph, which leads to HIB. The baseline is a hypergraph without information bottleneck of which the results are shown in the 1st row of Table~\ref{tab:ablation}. By introducing information bottleneck to this baseline, we can obtain the HIB of which performance is presented in the 2nd row of Table~\ref{tab:ablation}. Comparing the first two rows, we can see that the information bottleneck improves the AUC by 2.67 and NPV by 5.3, despite a decrease in PPV.  

\textbf{Importance of CHAD.} We then apply CHAD to HIB to verify its importance and show their results in Table~\ref{tab:ablation}. CHAD increases the AUC from 72.36 to 79.23, and PPV from 61.43 to 76.13, which is an impressive improvement. We owe it to the CHAD for discriminative feature extraction. 

\textbf{Necessity of CroC.} CroC aims to leverage unlabelled data to boost the performance. With CroC and unlabelled data used, the AUC, PPV, and NPV can be further improved over the HIB+CHAD. This observation highlights the necessity of effectively {utilising} unlabelled data. 

\begin{figure}[tb]
    \centering
    \includegraphics[trim=5 10 5 50, clip, width=0.65\linewidth]{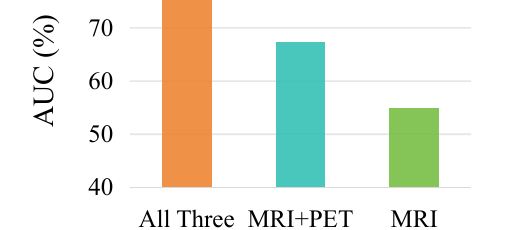}
    \caption{Ablation study on the modalities. All Three means using all three modalities, i.e., MRI, PET, and Non-imaging data.}
    \label{fig:ablation_modalities}
    \vspace{-0.4cm}
\end{figure}

\textbf{Efficacy of multi-modal data.}
{Since our \alg{} uses MRI, PET, and Non-image data, we evaluate the efficacy of our method in leveraging multi-modal data in Figure~\ref{fig:ablation_modalities}. The results show that \textbf{1)} using all three modalities achieves the best AUC of 80.79\%. \textbf{2)} Removing either the Non-Image (``All Three'' vs. ``MRI+PET'') or PET modality (``MRI+PET'' vs. ``MRI'') can decrease performance considerably, with both modalities having similar impacts on the performance. These observations demonstrate the importance of these modalities. 
}

{Overall, these findings demonstrate that \alg{} effectively integrates multi-modal data to enhance MCI progression prediction while offering a more efficient, clinically viable approach by {minimising} reliance on PET.
}

{\textbf{Ablation on augmentation policies.} To validate the effectiveness of the augmentation policies as stated in Section \ref{sec:data_preprocess}, we will first provide the rationale for using these augmentations and then present the experimental results to support our design of the augmentations.} 

{Firstly, horizontal flipping has been used in neuroimaging tasks~\cite{liu2023cascaded}, as it does not significantly alter the global anatomical features used in Alzheimer's classification. Random rotation and contrast adjustment serve to encourage the model to learn robust and invariant features from anatomical variations, acquisition settings, or noise. We restrict rotation angles to be small ($\pm$15$^{\circ}$) and contrast adjustments are bounded (with a gamma value in [0.6, 1.5]). This can either be diagnostically tolerable or simulate scanner-related variability.}

{Secondly, to validate these design choices, we perform ablation studies removing specific augmentations to assess their impact. We can see from Table~\ref{tab:ablation_augment} that removing any of them can decrease the performance.  This suggests that these augmentations can probably preserve diagnostic semantics.}

\begin{table}[tb]
    \centering
    \caption{{Ablation on the augmentation policies used in CroC for the task of MCI progression to AD vs. MCI vs. NC.}}
    \begin{tabular}{c|cccc}
     \toprule[1pt]
     \textbf{Settings} & Default aug. &  w/o flipping & w/o rotation & w/o contrast 
      \T \B \\
     \hline
     \T
      \textbf{AUC (\%)}  & \textbf{80.71} & 80.48  & 79.45 & 79.46\\
    \hline
    \toprule[1pt]
    \end{tabular}
    \label{tab:ablation_augment}
    \vspace{-0.6cm}
\end{table}

\subsubsection{{Sensitivity Analysis on Hyperparameters}}
\begin{figure*}[tb]
    \centering
    \includegraphics[trim=0 302 5 20, clip, width=0.9\textwidth]{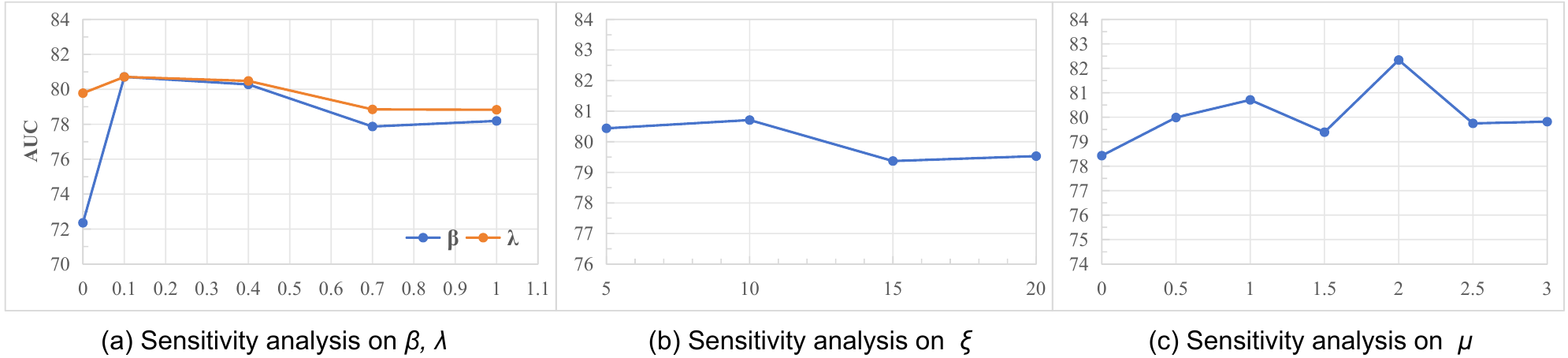}
    \vspace{-0.2cm}
    \caption{{Sensitivity analysis on hyperparameters, i.e., $\lambda$, $\beta$, $\xi$, and $\mu$, for the task of MCI progression to AD vs. MCI vs. NC.}}
    \label{fig:hyperparam}
    \vspace{-0.3cm}
\end{figure*}

{This subsection provides the sensitivity analysis on $\lambda$, $\beta$, $\xi$, and $\mu$. The first hyperparameter is in the CroC loss of Eq. (\ref{eq:croc_loss}), while the latter three are in Eq. (\ref{eq:loss_all}). The AUC (\%) of varying hyperparameters is shown in Fig.~\ref{fig:hyperparam}. We can see that the default settings of $\lambda=0.1$, $\beta=0.1$, $\xi=10$ can generally achieve the best AUC performance. However, the default $\mu=1$ is suboptimal with an AUC of 80.71\% while $\mu=2$ brings the best AUC of 82.34\%. This is an exciting boost, which suggests that the performance of our HIBMatch can be further improved when the hyperparameters are carefully tuned.}

\subsubsection{{Analysis of the Runtime and Memory Consumption}}
{We discuss the runtime, memory, and scalability as follows. 1) \textbf{Runtime}. We train \alg{} for about 7 hours, but it usually converges in 1 hour. When we stop training in 1 hour, the AUC is similar to that of 7-hour training, i.e., 80.58\% vs. 80.71\% (the first two rows of Table~\ref{tab:runtime_mem}). This comparison demonstrates that early stopping can save running time while keeping the performance almost the same. 
2) \textbf{Memory}. To make full use of our Nvidia A100-80GB GPU, we use 30 subjects (60 images, including both MRI and PET) to formulate a batch, costing about 73GB for training. However, such a batch size is usually large in the 3D case. If limited computational resources are available in real-world deployment, we can reduce the batch size to 8 as in ~\cite{pan2020multi,huang2019diagnosis}. This will cost about 20GB of memory, and the AUC performance only shows a subtle change, 80.69\% vs. 80.71\%. 
3) \textbf{Scalability}. The inference time and memory can be more important for real-world deployment than training, as only inference is involved in clinical use. The last row of Table~\ref{tab:runtime_mem} shows that the inference costs only 3.4GB and 22 seconds to test 48 subjects (i.e., about 2.18 subjects per second), which is efficient. The above analysis verifies the scalability of our method to accommodate different computational demands (e.g., runtime and memory).
}

\begin{table}[tb]
    \centering
    \caption{{Analysis on the runtime and memory consumption.}}
    \begin{tabular}{c|ccc}
     \toprule[1pt]
     \textbf{Methods} & \textbf{AUC (\%)} & \textbf{Runtime} & \textbf{Memory}
      \T \B \\
     \hline
     \T
       Default training setting  & 80.71 & 7 hours &73GB\\
        Training with early stopping &80.58 & 1 hour & 73GB\\
        Training with batch size of 8 & 80.69 & 7 hours &20GB\\
        Inference only & 80.71 &22 seconds & 3.4GB\\
    \hline
    \toprule[1pt]
    \end{tabular}
    \label{tab:runtime_mem}
    \vspace{-0.5cm}
\end{table}

\subsubsection{{Graph vs. Hypergraph}}
\begin{table}[tb]
    \centering
    \caption{{Comparison of hypergraph with graph for the task of MCI progression to AD vs. MCI vs. NC.}}
    \begin{tabular}{c|ccc}
     \toprule[1pt]
     \textbf{Methods} & \textbf{AUC (\%)} & \textbf{PPV (\%)} & \textbf{NPV (\%)}
      \T \B \\
     \hline
     \T
       Hypergraph (Ours)  & 80.71 & 76.76 & 83.13\\
        Graph &78.86 & 72.15 & 80.48\\
    \hline
    \toprule[1pt]
    \end{tabular}
    \label{tab:graph}
    \vspace{-0.6cm}
\end{table}

{To justify the superiority of a hypergraph over a standard graph, we replace the hypergraph with a graph while keeping the other settings unchanged. Their results are compared in the Table~\ref{tab:graph}. We can see that the graph is worse than the hypergraph version in AUC, which demonstrates the superiority of the hypergraph in HIBMatch.}

\subsubsection{Visualization}
\begin{figure*}[tb]
    \centering
    \includegraphics[trim=15 100 45 80, clip, width=0.83\textwidth]{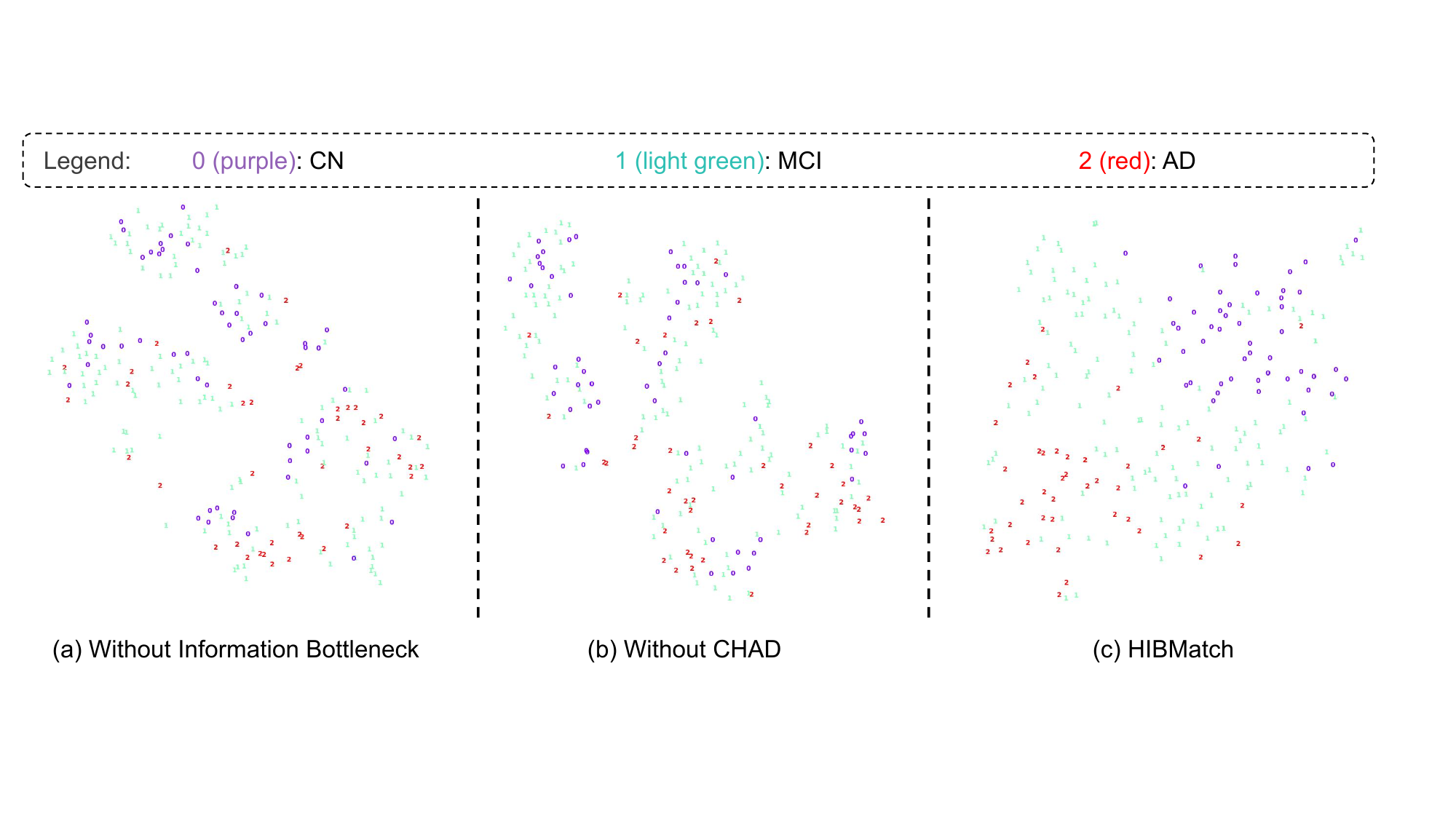}
    \vspace{-0.2cm}
    \caption{{Visualization on feature distributions. (a) Removing the information bottleneck from HIBMatch; (b) Removing CHAD from HIBMatch; (c) HIBMatch. Different digits denote: ``0'' for MCI progression to CN, ``1'' for MCI (status unchanged), and ``2'' for MCI progression to AD (MCIc).}}
    \label{fig:visualize_feat}
    \vspace{-0.4cm}
\end{figure*}


{
To support the effectiveness of Hypergraph Information Bottleneck (HIB) and the Consistency between HIB and a Discriminative Classifier (CHAD), we include ablation results in Fig.~\ref{fig:visualize_feat}(a) and Fig.~\ref{fig:visualize_feat}(b), where removing either component leads to greater feature overlap—particularly between CN and AD—demonstrating their contribution to more discriminative embedding learning. It is worth noting that our goal is to promote more discriminative feature learning through the use of HIB and CHAD, not to guarantee perfect class separation. While complete separation is not achieved (as reflected in the non-perfect classification performance), the inclusion of both HIB and CHAD consistently improves class distinction in the embedding space.
}

\subsubsection{Evaluation on Independent Test Set}
We further evaluate our \alg{} on an independent test set where the subjects are unseen during training. The independent test set has 57 subjects (4, 46, and 7 for NC, MCI, and AD, respectively) in addition to the 508 subjects in the training set.  We can see from Table~\ref{tab:test_set_results} that \alg{} can still obtain {favourable} results on the test set when predicting MCI progression to AD vs. NC (i.e., MCIc vs. NC of MCInc) and MCI progression to AD vs. MCI (i.e., MCIc vs. MCI of MCInc). These promising results on the unseen test set suggest that \alg{} has the potential to be deployed effectively in new clinical environments.


\subsubsection{{Evaluation of Generalisation Ability on AIBL}}
{Finally, we validate the generalisation ability of our HIBMatch on AIBL by training a model on ADNI and evaluating it on AIBL. We introduce the experimental settings and the results below.}

{
We only use MRI and non-imaging data to retrain a new model on ADNI which will be tested on AIBL using the same two modalities. Only these two modalities are used because AIBL does not have subjects meeting the following three criteria: 1) diagnosed as MCI at the baseline visit, 2) with diagnosis results two/three years later, 3) having complete three modalities (MRI, PET, and non-imaging data). 
We thus discard PET to have sufficient subjects from AIBL for testing and only leverage MRI and non-imaging modalities to train the HIBMatch model on ADNI. The non-imaging information includes age, gender, and MMSE because the other information, such as ADNI-MEM and ADNI-EF, is unavailable in AIBL. After filtering, we identify 391 MCI subjects from ADNI who have both MRI and non-imaging data at baseline and corresponding diagnostic outcomes three years later. An additional 38 subjects without follow-up diagnoses are treated as unlabelled data. We split the subjects into five folds, with four folds for training and the remaining one for validation. This facilitates five-fold cross-validation. The model achieving the best validation performance in each fold will be evaluated on AIBL, which has 119 eligible subjects. Since no MCI subjects in AIBL revert to cognitive normal status, only two outcome classes are considered. The same filtering strategy is applied to both datasets. Table~\ref{tab:demographic_aibl} shows the demographics of the included subjects used in these two datasets.
}

\begin{table} [!t]
    \centering
    \caption{{Demographics of the included subjects in ADNI and AIBL.}}
    \label{tab:demographic_aibl}
    \setlength{\tabcolsep}{0.5mm}{
    \begin{tabular}{l|cc|c|cc}
        \toprule[1pt]
        \textbf{Datasets} & \multicolumn{3}{c|}{\textbf{ADNI (Train/Val)}} & \multicolumn{2}{c}{\textbf{AIBL (Test)}}\\
        \hline
        Label type &  \multicolumn{2}{c|}{\textbf{Labelled}} & \multirow{2}{*}{\textbf{Unlabelled}} &\multicolumn{2}{c}{\textbf{Labelled}}\\
        Progression to & MCI & AD &  & MCI & AD\\
        \toprule[1pt]
        Number       & 295 & 96 & 38 & 112 &7\\
        Gender (M/F) & 180/115 & 52/44 & 15/23 &53/59 & 5/2 \\
        Age          &71.67$\pm$7.33 &72.50$\pm$6.71 & 69.38$\pm$7.25 & 70.42$\pm$9.40 & 68.71$\pm$6.67\\
        \toprule[1pt]
    \end{tabular}
    }
    \vspace{-0.4cm}
\end{table}

{
We train the model for 50 epochs using the default setting as stated in Section~\ref{sec:training_details}. The last row of Table~\ref{tab:test_set_results} shows that the AUC of our \alg{} on the AIBL does not show a clear decrease, and the accuracy is still high, compared with the performance on the independent test set. The experimental results verify the promising generalisation ability of \alg{}. 
}

\begin{table}[tb]
    \centering
    \caption{Evaluation of \alg{} on an independent test set {and evaluation on the generalisation ability of HIBMatch on AIBL}. The independent test set has two tasks: MCI progression to  1) AD (MCIc) vs. MCI progression to NC, and 2) AD (MCIc) vs. MCI unchanged over time, while the generalisation experiment implements the latter only. The subjects in the test set are unseen during training.}
    \begin{tabular}{c|c|cc}
    \hline
    \textbf{Tasks} & \textbf{Splits}& \textbf{AUC} & \textbf{ACC} \\
    \hline
       \multirow{2}{*}{AD (MCIc) vs. NC} &Validation  & 95.23 &94.59 \\
        &Independent Test  &92.86 &89.27 \\
        \hline
       \multirow{2}{*}{AD (MCIc) vs. MCI} & Validation & 85.93 &87.68 \\
        &Independent Test & 80.16 &81.50\\
    \hline
    \hline
    {AD (MCIc) vs. MCI} & {Test on AIBL} & 78.67 & 94.29\\
    \hline
    \end{tabular}
    \label{tab:test_set_results}
    \vspace{-0.4cm}
\end{table}

\subsubsection{Limitations}
A major limitation of \alg{} is that it requires all three modalities for training and inference to ensure high performance. However, acquiring all MRI, PET, and Non-imaging data can be expensive in clinical practice, sometimes even unavailable. Therefore, how to effectively {utilise} incomplete modalities for better MCI progression prediction can be meaningful for future work.

\section{Conclusion}
\label{conclusion}

Our proposed \alg{} demonstrated promising results in addressing the challenges associated with Alzheimer's disease prognosis leveraging multimodal data. \alg{} {utilises} a Hypergraph Information Bottleneck to extract discriminative features from multimodal data, with the help of hypergraphs and the principle of information bottleneck. The discriminative features are then enhanced by a consistency {regularisation} on the hypergraph and a discriminative classifier. \alg{} further includes a cross-modal contrastive loss to make full use of unlabelled data for better performance. Our proposed method outperforms other state-of-the-art methods in semi-supervised learning settings and under fewer annotations. Our results also demonstrated better robustness of \alg{} under both topological and feature perturbations. The success of \alg{} highlights the potential of hypergraph-based methods in the field of Alzheimer's disease prognosis prediction and sets a foundation for future research in this area.



\section*{Acknowledgment}
Data used in the preparation of this article were obtained from the Alzheimer’s Disease Neuroimaging Initiative (ADNI) database (adni.loni.usc.edu). As such, the investigators within the ADNI contributed to the design and implementation of ADNI and/or provided data but did not participate in the analysis or writing of this report. A complete listing of ADNI investigators can be found at: \url{https://adni.loni.usc.edu/wp-content/uploads/how_to_apply/ADNI_Acknowledgement_List.pdf}

This project was supported with funding from the Cambridge Centre for Data-Driven Discovery and Accelerate Program for Scientific Discovery, made possible by a donation from Schmidt Sciences. This work was also partially supported by RGC Collaborative Research Fund (No. C5055-24G), the Start-up Fund of The Hong Kong Polytechnic University (No. P0045999), the Seed Fund of the Research Institute for Smart Ageing (No. P0050946), and Tsinghua-PolyU Joint Research Initiative Fund (No. P0056509), and PolyU UGC funding (No. P0053716). 
Other grants that support this work include: 
 ZD from Accelerate-C2D3 Funding, University of Cambridge and Wellcome Trust 221633/Z/20/Z; SW from Wellcome Trust 221633/Z/20/Z; 
 AIAR from CMIH and CCIMI, University of Cambridge;
 ZK acknowledges support from the Biotechnology and Biological Sciences Research Council H012508 and BB/P021255/1, Alan Turing Institute TU/B/000095, Wellcome Trust 205067/Z/16/Z, 221633/Z/20/Z, Royal Society INF/R2/202107;
 CBS from the Philip Leverhulme Prize, the Royal Society Wolfson Fellowship, the EPSRC advanced career fellowship EP/V029428/1, EPSRC grants EP/S026045/1 and EP/T003553/1, EP/N014588/1, EP/T017961/1, the Wellcome Innovator Awards 215733/Z/19/Z and 221633/Z/20/Z, the European Union Horizon 2020 research and innovation program under the Marie Skodowska-Curie grant agreement No. 777826 NoMADS, the Cantab Capital Institute for the Mathematics of Information and the Alan Turing Institute.

We also thank Prof. Eleni Vasilaki, Dr. Yuanxi Li, and Stefanos Ioannou for their insightful discussion.

\section*{References}
\bibliographystyle{IEEEtran}

\bibliography{refs}

\end{document}